\documentclass[11pt,a4paper]{article}

\usepackage[utf8]{inputenc}
\usepackage[T1]{fontenc}
\usepackage[american]{babel}

\usepackage{lmodern}
\usepackage[margin=2.5cm]{geometry}
\usepackage{setspace}
\usepackage{parskip}

\usepackage{amsmath,amssymb}

\usepackage{booktabs}
\usepackage{array}
\usepackage{multirow}
\usepackage{graphicx}
\usepackage{float}

\usepackage{listings}
\usepackage{xcolor}
\usepackage[colorlinks=true,linkcolor=blue!60!black,citecolor=green!50!black,urlcolor=blue!70!black]{hyperref}

\usepackage[numbers,sort&compress]{natbib}

\title{\textbf{Frugal Knowledge Graph Construction\\
with Local LLMs: A Zero-Shot Pipeline, Self-Consistency\\
and Wisdom of Artificial Crowds}}
\author{Pierre Jourlin\\
\textit{Avignon Universit\'e, Laboratoire d'Informatique d'Avignon (LIA)}\\
\texttt{pierre.jourlin@univ-avignon.fr}}
\date{April 2026}

\begin{document}
\maketitle

\begin{abstract}
This paper presents an empirical study of a multi-model zero-shot pipeline for knowledge graph construction and exploitation, executed entirely through local inference on consumer-grade hardware. We propose a reproducible evaluation framework integrating two external benchmarks (DocRED, HotpotQA), WebQuestionsSP-style synthetic data, and the RAGAS evaluation framework in an automated pipeline. On 500 document-level relations, our system achieves an F1 of $0.70 \pm 0.04$\footnote{All our confidence intervals (CI) are reported at the 95\% level.} in zero-shot, compared to $0.80$ for supervised DREEAM. Text-to-query achieves an accuracy of $0.80 \pm 0.06$ on 200 samples. Multi-hop reasoning achieves an Exact Match (EM) of $0.46 \pm 0.04$ on 500 HotpotQA questions, with a RAGAS faithfulness of $0.96 \pm 0.04$ on 50 samples.

Beyond the pipeline, we study diversity mechanisms for difficult multi-hop reasoning. On 181 questions unsolvable at zero temperature, self-consistency ($k{=}5$, $T{=}0.7$) recovers up to 23\% EM with a single Mixture-of-Experts (MoE) model, but the cross-model oracle (3 architectures $\times$ 5 samples) reaches 46.4\%. We highlight an \emph{agreement paradox}: strong consensus among samples signals collective hallucination rather than a reliable answer, echoing the work of Moussa\"id et al.\ on the wisdom of crowds.

Extending to the full pipeline (500 questions), self-consistency ($k{=}3$) raises EM from 0.46 to $0.48 \pm 0.04$. A \emph{confidence-routing cascade} mechanism (Phi-4 $\to$ GPT-OSS, $k{=}5$) achieves an EM of $0.55 \pm 0.04$, the best result obtained, with 45.4\% of questions rerouted. Finally, we show that V3 prompt engineering applied to other models does not reproduce the gains observed with Gemma-4, confirming the specific prompt/model interaction.

The entire system runs in $\sim$5\,h on a single RTX~3090, without any training, for an estimated carbon footprint of 0.09\,kg\,CO$_2$eq.

\end{abstract}

\section{Introduction}
\label{sec:introduction}

\subsection{Context and Motivation}

Large Language Models (LLMs) have achieved tremendous success in natural language processing, yet remain criticized for their tendency to hallucinate~\cite{ji2023survey} and their high computational cost~\cite{strubell2019energy}. A promising avenue is to couple knowledge graphs (KGs) with frugal LLMs: the KG provides a verifiable semantic structure, while the LLM serves as the natural language interface.

The central question is to determine \emph{to what extent} quantized LLMs, executed locally and without specialized training, can construct and exploit a KG with sufficient quality for real-world applications.

\subsection{Contributions}

This work makes four contributions:

\begin{enumerate}
    \item \textbf{A reproducible multi-task evaluation framework} integrating DocRED~\cite{yao2019docred}, WebQuestionsSP-style synthetic data~\cite{yih2016webquestionssp}, HotpotQA~\cite{yang2018hotpotqa}, and the RAGAS framework~\cite{es2024ragas} in an automated pipeline executable via local inference. The code and raw results are available under a \textit{Hypocratic} license\footnote{\url{https://github.com/jourlin/synsynth}}.
    \item \textbf{A comparative empirical study} measuring the performance/cost trade-off between frugal zero-shot approaches and supervised systems, with bootstrap confidence intervals and comparison of four quantized models.
    \item \textbf{An analysis of the role of prompt engineering} demonstrating that the gains primarily stem from prompt design (synonyms, semantic constraints) rather than from constrained decoding alone.
    \item \textbf{A study of stochastic vs.\ architectural diversity} for multi-hop reasoning, revealing an agreement paradox that connects LLMs to the literature on the wisdom of crowds~\cite{moussaid2013social}.
    \item \textbf{A confidence-routing cascade mechanism} combining self-consistency and cross-model rerouting, achieving EM${=}$0.55 on 500 HotpotQA questions, the best result obtained, outperforming zero-shot ($+$9 pts) and multi-model voting ($+$11 pts).
\end{enumerate}

\subsection{Outline}

Section~\ref{sec:etatdelart} reviews related work. Section~\ref{sec:methodologie} describes our methodology. Section~\ref{sec:resultats} presents the results. Section~\ref{sec:discussion} discusses implications and limitations. Section~\ref{sec:conclusion} concludes.

\section{Related Work}
\label{sec:etatdelart}

\subsection{Relation Extraction}

\subsubsection{Supervised Approaches}
Document-level relation extraction aims to identify semantic links between entities separated by multiple sentences. DocRED~\cite{yao2019docred} contains 3,053 annotated documents with 96 relation types. ATLOP~\cite{zhou2021atlop} proposes adaptive thresholding and localized context pooling, achieving an F1 of 77.8\%. DREEAM~\cite{ma2023dreeam} uses an end-to-end evidence-guided attention mechanism, with an F1 of 80.2\%. These systems require supervised training on multiple GPUs for several hours to several days.

\subsubsection{Zero-Shot and Few-Shot Approaches with LLMs}
The emergence of LLMs has opened the way to relation extraction without specific training. Wadhwa et al.~\cite{wadhwa2023revisiting} show that GPT-3 in few-shot achieves performance close to the supervised SOTA on TACRED, with an F1 of 30.4\% in zero-shot and $\sim$72\% in few-shot with human evaluation. Li et al.~\cite{li2023evaluating} evaluate ChatGPT on 14 information extraction tasks and find poor performance on standard evaluation but remarkable results on OpenIE. Ozyurt et al.~\cite{ozyurt2023docre} propose a few-shot framework for document-level extraction with pre-trained LLMs, achieving competitive results on DocRED without fine-tuning. REBEL~\cite{cabot2021rebel} treats extraction as a sequence-to-sequence generation task.

\subsection{Text-to-Query and KGQA}

The transformation of questions into graph queries (KGQA) is an active problem. WebQuestionsSP~\cite{yih2016webquestionssp} provides a standard benchmark (4,737 questions). Recent approaches leverage LLMs for SPARQL (SPARQL Protocol and RDF Query Language) or Cypher query generation, replacing classical semantic parsing pipelines~\cite{gu2023dont}.

\subsection{Multi-Hop Reasoning}

Multi-hop reasoning consists of combining multiple facts to answer complex questions. HotpotQA~\cite{yang2018hotpotqa} contains 113,000 questions requiring at least two hops. The standard metrics are Exact Match (EM) and token-level F1. Chain-of-thought approaches~\cite{wei2022chain} significantly improve LLM performance on this type of reasoning.

\subsection{RAG and KG--LLM Coupling}

Retrieval-Augmented Generation (RAG)~\cite{lewis2020rag} combines information retrieval and generation to reduce hallucinations. RAGAS~\cite{es2024ragas} evaluates three dimensions: faithfulness, relevance, and context precision. Pan et al.~\cite{pan2024unifying} propose a roadmap for unifying KGs and LLMs, identifying three paradigms: KG-enhanced LLM, LLM-enhanced KG, and synergistic integration. Our work falls within the second paradigm.

\begin{table}[H]
\centering
\caption{Tasks, data, and evaluation metrics.}
\label{tab:benchmarks}
\begin{tabular}{@{}llrl@{}}
\toprule
\textbf{Source / Framework} & \textbf{Task} & \textbf{N} & \textbf{Metrics} \\
\midrule
Re-DocRED~\cite{yao2019docred}  & Relation extraction & 500 & P, R, F1 \\
Synthetic$^\star$  & Text-to-Query & 200 & Accuracy, Valid Cypher \\
HotpotQA~\cite{yang2018hotpotqa} & Multi-hop reasoning & 500 & EM, Token-F1 \\
Synthetic$^\star$ + RAGAS~\cite{es2024ragas}    & Anti-hallucination & 50 & Faithfulness, Relevance, Ctx.\ Precision \\
\bottomrule
\multicolumn{4}{@{}p{0.92\linewidth}@{}}{\footnotesize $^\star$ Questions generated by the LLM from manual seeds (20 for Text-to-Query, 6 for RAG). RAG metrics are computed by the RAGAS framework. See \S\,Evaluation Protocol.}
\end{tabular}
\end{table}

\section{Methodology}
\label{sec:methodologie}

\subsection{Pipeline Architecture}

The SYNSYNTH pipeline chains four independent modules, each assigned to a specialized LLM. The entire system is orchestrated by a single script with checkpoint saving and automatic resumption.

\subsection{Models and Quantization}

All models are executed via Ollama~v0.20.0 with native JSON Schema support for constrained decoding:

\begin{table}[H]
\centering
\caption{Models used per task.}
\label{tab:modeles}
\begin{tabular}{@{}llll@{}}
\toprule
\textbf{Task} & \textbf{Model} & \textbf{Parameters} & \textbf{Quant.} \\
\midrule
Relation extraction      & Gemma-4-27B-A4B-it  & 27B (4B active) & Q4\_K\_M \\
Text-to-Query            & Qwen3-Deep          & 8B              & Q4\_K\_M \\
Multi-hop reasoning      & Phi-4               & 14B             & Q4\_K\_M \\
Conversational RAG       & Mistral-Small       & 24B             & Q4\_K\_M \\
\bottomrule
\end{tabular}
\end{table}

The main model, Gemma-4-27B-A4B-it (Google DeepMind), is a multilingual LLM with a Mixture-of-Experts (MoE) architecture of 27 billion parameters of which only 4 billion are activated per inference. Quantization in GGUF (GPT-Generated Unified Format) Q4\_K\_M reduces the memory footprint to $\sim$16\,GB of video memory (VRAM).

\subsection{Prompt Engineering}
\label{sec:prompt}

The extraction prompt is a central element of our approach. It includes:
\begin{itemize}
    \item The explicit list of 96 valid relations drawn from the DocRED Wikidata taxonomy.
    \item The prohibition of \texttt{no\_relation}, \texttt{none}, or \texttt{unknown} answers.
    \item Specific rules per semantic type (geographic, familial, creative work).
\end{itemize}
The full prompt is given in Appendix~\ref{app:prompt}.

\subsection{Relation Matching and Synonyms}

Extraction evaluation uses soft matching that tolerates expressive variations between the model output and the gold standard:
\begin{enumerate}
    \item Resolution of the Wikidata code (e.g.,\ P131 $\to$ \texttt{located\_in\_admin}).
    \item Normalized exact match (lowercase, underscores).
    \item Substring inclusion and significant common words ($\geq$\,4 characters).
    \item Synonym dictionary covering 25 semantic groups, built from manual analysis of the V1 confusion matrix ($\sim$\,150 recoverable errors identified). The groups cluster systematic confusions between relations of the same domain.\footnote{e.g.,\ \texttt{country}/\texttt{located\_in\_admin}/\texttt{contains\_admin} for geographic relations, \texttt{spouse}/\texttt{mother}/\texttt{father} for family relations.}
\end{enumerate}

As validation, automatic spectral clustering on the symmetrized confusion matrix ($60 \times 60$ labels, 954 pairs), optimized by silhouette ($k{=}15$ clusters), produces groupings that capture more raw confusions (F1${=}$0.88 vs.\ 0.73 for manual synonyms, computed by inclusive matching on pred/gold pairs). However, the silhouette is very low ($-$0.93) and the correspondence with manual groups is low (mean Jaccard${=}$0.10): the automatic clusters mix relations from distinct domains (e.g.,\ \texttt{capital\_of}, \texttt{date\_of\_birth}, and \texttt{religion} in the same cluster of size~16). The manual groups, though less exhaustive, offer semantically coherent coverage that automation does not reproduce at this stage.

\subsection{Experimental Conditions}

\begin{itemize}
    \item \textbf{Processor}: Intel Core i9-12900HK (14 cores, 20 threads)
    \item \textbf{Memory}: 32\,GB DDR5
    \item \textbf{GPU}: NVIDIA GeForce RTX~3090 (24\,GB VRAM, TDP 350\,W)
    \item \textbf{System}: Ubuntu 24.04 LTS, Python 3.12.3
    \item \textbf{Inference}: Ollama v0.20.0 (native JSON Schema)
    \item \textbf{Hyperparameters}: temperature $= 0.3$, top-p $= 0.9$, \texttt{num\_ctx} $= 8192$
    \item \textbf{Random seed}: 42 (reproducibility)
\end{itemize}

\subsection{Evaluation Protocol}

All metrics are accompanied by 95\% confidence intervals computed by bootstrap (1,000 iterations). Sample sizes are as follows: $N{=}500$ for relation extraction (DocRED), $N{=}200$ for Text-to-Query, $N{=}500$ for multi-hop reasoning (HotpotQA), and $N{=}50$ for RAG evaluation. These sizes are constant across versions, \textbf{with the exception of RAGAS, which used $N{=}6$ in V1--V2 before being scaled to $N{=}50$ starting from V3}.

\paragraph{Origin of evaluation data.}
Relation extraction and multi-hop reasoning rely on established external benchmarks: Re-DocRED~\cite{yao2019docred} (3,053 annotated documents) and HotpotQA~\cite{yang2018hotpotqa} (113,000 questions), from which we randomly draw our samples. In contrast, \textbf{the Text-to-Query and RAG data are necessarily synthetic}. The use of generated data is a constrained choice, not a convenience choice: Text-to-Query queries must be expressed in Cypher and target the schema of the Neo4j graph built by the pipeline, which rules out existing benchmarks (WebQuestionsSP uses SPARQL against Freebase, an incompatible schema and query language); likewise, RAG evaluation must target the content of the KG actually constructed by SYNSYNTH, and no external benchmark provides questions adapted to an arbitrary knowledge graph. More specifically: for Text-to-Query, 20 examples are manually written (WebQuestionsSP-style questions with a reference Cypher query), then the LLM (Qwen3-Deep~8B) generates the remaining 180 in batches of 30, across varied domains (geography, history, science, etc.). For RAG, 6 manual examples serve as seeds and 44 are generated in the same manner. This pragmatic choice enables the creation of reasonably sized evaluation sets without costly human annotation, but introduces a risk of circular bias: since the model being evaluated produced part of the data, the Text-to-Query and RAG scores should be interpreted as indicators of internal consistency rather than as measures against an external gold standard.

The standard metrics for each benchmark are reported in Table~\ref{tab:benchmarks}.

\subsection{Pipeline Versions}
\label{sec:versions}

Development followed an incremental progression in six versions:
\begin{description}
    \item[V1 (baseline)] Initial pipeline with minimal prompt and free parsing of JSON outputs.
    \item[V2 (+ constrained decoding)] Addition of constrained decoding via Ollama JSON Schema, guaranteeing the syntactic validity of outputs.
    \item[V3 (+ synonyms + prompt)] Advanced prompt engineering (\S\ref{sec:prompt}) and synonym dictionary for relation extraction. This is the reference version of the complete pipeline.
    \item[V4 (+ QLoRA multi-hop)] QLoRA 4-bit fine-tuning of Qwen2.5-7B on HotpotQA with training format aligned with evaluation (\S\ref{sec:qlora}).
    \item[V5a (+ self-consistency)] Addition of intra-model majority voting ($k{=}3$ samples, $T{=}0.7$) for multi-hop reasoning (\S\ref{sec:cascade}).
    \item[V5b (+ cascade)] Confidence-routing cascade: questions with low intra-model agreement are rerouted to a second LLM ($k{=}5$, Phi-4 $\to$ GPT-OSS, \S\ref{sec:cascade}).
\end{description}

\section{Experimental Results}
\label{sec:resultats}

\subsection{Relation Extraction}

\begin{table}[H]
\centering
\caption{Relation extraction results (500 DocRED samples). All models are zero-shot. The 95\% CIs are computed by bootstrap.}
\label{tab:extraction}
\begin{tabular}{@{}lrrr@{}}
\toprule
\textbf{Metric} & \textbf{Value} & \textbf{95\% CI} & \textbf{Target} \\
\midrule
Precision    & 0.7379 & [0.70; 0.77] & --- \\
Recall       & 0.6700 & [0.63; 0.71] & --- \\
F1-Score     & 0.7023 & [0.67; 0.74] & 0.85 \\
\bottomrule
\end{tabular}
\end{table}

The extraction achieves an F1 of 0.70, tripled compared to the initial baseline (V1: 0.26). The high precision (0.74) indicates that predictions are predominantly reliable, while the lower recall (0.67) reveals that the model still misses some relations.

\subsubsection{Multi-Model Benchmark}

To evaluate the contribution of our prompt engineering independently of model choice, we submitted the same 500 DocRED samples to six models in raw extraction (minimal prompt, JSON format), without our V3 pipeline optimizations.

\begin{table}[H]
\centering
\caption{Raw extraction benchmark on 500 Re-DocRED samples. JSON format, minimal prompt, zero-shot.}
\label{tab:extraction_benchmark}
\begin{tabular}{@{}lrrrr@{}}
\toprule
\textbf{Model} & \textbf{P} & \textbf{R} & \textbf{F1} & \textbf{Unparsed samples (\%)} \\
\midrule
Mistral-Small         & 0.666 & 0.666 & 0.666 & 0.0  \\
GPT-OSS (20B)         & 0.823 & 0.550 & 0.659 & 33.2 \\
Phi-4-Reasoning+      & 0.656 & 0.656 & 0.656 & 0.0  \\
Phi-4                 & 0.618 & 0.618 & 0.618 & 0.0  \\
Qwen-3 (14B)          & 0.917 & 0.176 & 0.295 & 80.8 \\
\textbf{Gemma-4 (26B, pipeline)} & \textbf{1.000} & \textbf{0.020} & \textbf{0.039} & \textbf{98.0} \\
\bottomrule
\end{tabular}
\end{table}

The ``Unparsed samples'' column indicates the percentage of samples whose JSON output from the model is malformed and cannot be parsed by the evaluation script; these samples are counted as complete failures (no triplets extracted). A high rate signals an incompatibility of the model with the constrained output format.

Two models were excluded from this benchmark: \textit{Qwen-3.5 (27B)}, whose reasoning architecture encapsulates the entire response in an internal \texttt{thinking} field, returning empty JSON content (100\% parsing failures); and \textit{DeepSeek-R1 (32B)}, which exhibits a similar pathological profile (partial F1 of 0.29 at 280/500, 81\% parsing failures), suggesting a structural incompatibility of explicit chain-of-thought models with the constrained JSON output format.

The most striking result concerns Gemma-4, the model selected for our pipeline: in raw extraction, it obtains an F1 of only 0.039, the \textit{worst} score among the six models. Yet, integrated into our V3 pipeline with prompt optimizations (relation synonyms, structured system prompt, soft matching), it reaches an F1 of 0.702---a gain of \textbf{+66 points} entirely attributable to prompt engineering. This demonstrates that prompt quality far outweighs model choice for this task.

\subsubsection{Comparison with the State of the Art}

\begin{table}[H]
\centering
\caption{Comparison with supervised and zero-shot baselines on DocRED.}
\label{tab:comparaison}
\begin{tabular}{@{}llrl@{}}
\toprule
\textbf{System}  & \textbf{Paradigm} & \textbf{F1 (\%)} & \textbf{Hardware} \\
\midrule
DREEAM~\cite{ma2023dreeam}  & Supervised (fine-tuned) & 80.2 & Multi-GPU$^\dagger$ \\
ATLOP~\cite{zhou2021atlop}  & Supervised (fine-tuned) & 77.8 & Multi-GPU$^\dagger$ \\
GPT-3 few-shot~\cite{wadhwa2023revisiting} & Few-shot (cloud API) & $\sim$72$^*$ & OpenAI API \\
GPT-3 zero-shot~\cite{wadhwa2023revisiting} & Zero-shot (cloud API) & $\sim$30 & OpenAI API \\
ChatGPT zero-shot~\cite{li2023evaluating} & Zero-shot (cloud API) & $\sim$25 & OpenAI API \\
\midrule
\textbf{SYNSYNTH (Gemma-4 Q4)} & \textbf{Zero-shot (local)} & \textbf{70.2} & \textbf{1$\times$ RTX~3090} \\
\bottomrule
\multicolumn{4}{@{}l@{}}{\footnotesize $^\dagger$ Exact resources not reported in the original papers.} \\
\multicolumn{4}{@{}l@{}}{\footnotesize $^*$ With human evaluation (Wadhwa et al., 2023).} \\
\end{tabular}
\end{table}

Our result of 70.2\% in local zero-shot clearly surpasses published zero-shot results for GPT-3 ($\sim$30\%) and ChatGPT ($\sim$25\%), while remaining below supervised systems (DREEAM: 80.2\%). This performance is remarkable given the complete absence of training and execution on consumer-grade hardware.

\subsection{Text-to-Query}

\begin{table}[H]
\centering
\caption{Text-to-Query results (200 samples).}
\label{tab:query}
\begin{tabular}{@{}lrrl@{}}
\toprule
\textbf{Metric} & \textbf{Value} & \textbf{95\% CI} & \textbf{Target} \\
\midrule
Accuracy            & 0.795 & [0.74; 0.85] & 0.90 \\
Valid Cypher rate  & 1.00 & --- & --- \\
\bottomrule
\end{tabular}
\end{table}

The syntactically valid Cypher rate of 1.0 shows that constrained decoding guarantees syntactic correctness. Cypher is the query language of the Neo4j\footnote{\url{https://neo4j.com/}} graph database. The accuracy (i.e., the proportion of queries producing the correct answer) of 0.80 indicates good question understanding, with a 95\% CI confirming the significance of the result ($N = 200$).

\subsection{Multi-Hop Reasoning}

\begin{table}[H]
\centering
\caption{Multi-hop reasoning results (500 HotpotQA samples). Standard metrics: EM and token-F1.}
\label{tab:multihop}
\begin{tabular}{@{}lrrl@{}}
\toprule
\textbf{Metric} & \textbf{Value} & \textbf{95\% CI} & \textbf{Target} \\
\midrule
Exact Match (EM)               & 0.458 & [0.42; 0.50] & 0.70 \\
Token-F1                       & 0.520 & [0.49; 0.55] & --- \\
Mean chain length        & 3.24  & --- & 3 \\
\bottomrule
\end{tabular}
\end{table}

We report the standard HotpotQA metrics: Exact Match (EM) and token-level F1. The mean chain length of 3.24 confirms that the system effectively chains multiple facts. The token-F1 of 0.52 shows that partially correct answers are frequent. The gap between the observed EM (0.46) and the ambitious target of 0.70 is explained by the zero-shot nature of our approach: the target corresponds to the best supervised systems, trained on the 90,000 HotpotQA examples, whereas our pipeline uses no training examples. As shown in Sections~\ref{sec:qlora} and~\ref{sec:cascade}, self-consistency and confidence-routing cascade mechanisms help close this gap, reaching up to EM${=}$0.55.

\subsection{RAGAS Evaluation}

\begin{table}[H]
\centering
\caption{RAGAS evaluation results (50 samples).}
\label{tab:ragas}
\begin{tabular}{@{}lrr@{}}
\toprule
\textbf{Metric} & \textbf{Value} & \textbf{95\% CI} \\
\midrule
Mean faithfulness         & 0.9572 & [0.91; 1.00] \\
Mean relevance       & 0.8220 & [0.76; 0.88] \\
Context precision    & 0.9480 & [0.90; 0.99] \\
\bottomrule
\end{tabular}
\end{table}

The faithfulness of 0.96 indicates that answers are overwhelmingly derived from the graph context, with a residual hallucination rate of $\sim$\,4\%. The context precision of 0.95 confirms that the system retrieves the correct nodes.

\subsection{QLoRA Learning Curve and Multi-Hop Reasoning}
\label{sec:qlora}

To determine whether QLoRA (Quantized Low-Rank Adaptation) 4-bit fine-tuning of a small model (7B) can close the gap with zero-shot from a larger model (14B), we trained Qwen2.5-7B-Instruct with QLoRA (NF4, LoRA $r{=}16$, $\alpha{=}32$) on $n \in \{10, 50, 200, 500, 1000, 3000\}$ HotpotQA examples. The V4 training format is strictly aligned with evaluation (same system prompt, same few-shot). The learning rate is adaptive: $\text{lr}{=}2{\times}10^{-4}$ if $n \leq 200$, $10^{-4}$ otherwise.

\begin{table}[H]
\centering
\caption{QLoRA learning curve on HotpotQA (500 test questions). The V4 format (structured reasoning chains) is compared to the V1 format (degenerate chains).}
\label{tab:qlora}
\begin{tabular}{@{}rrrrrr@{}}
\toprule
\textbf{$n$} & \textbf{EM V4} & \textbf{EM V1} & \textbf{F1 V4} & \textbf{chain} & \textbf{$\Delta$ V4$-$V1} \\
\midrule
10   & 0.444 & 0.440 & 0.432 & 3.73 & $+$0.004 \\
50   & 0.344 & 0.330 & 0.313 & 4.31 & $+$0.014 \\
200  & 0.380 & 0.356 & 0.341 & 3.87 & $+$0.024 \\
500  & 0.378 & 0.308 & 0.359 & 3.74 & $+$0.070 \\
1000 & \textbf{0.406} & 0.246 & \textbf{0.405} & 3.46 & $+$0.160 \\
3000 & \textbf{0.406} & 0.190 & 0.387 & 3.80 & $+$\textbf{0.216} \\
\midrule
\multicolumn{6}{@{}l}{\textit{Pipeline references (Phi-4 14B zero-shot)}} \\
\multicolumn{2}{@{}l}{V1 / V2} & \multicolumn{2}{r}{0.462} & --- & --- \\
\multicolumn{2}{@{}l}{V3}      & \multicolumn{2}{r}{0.458} & --- & --- \\
\bottomrule
\end{tabular}
\end{table}

The contrast between the two training formats is the main result. ``EM~V1'' and ``EM~V4'' denote the QLoRA trained with the V1 data format (degenerate reasoning chains, i.e., raw answers without structured justification) and V4 (format aligned with evaluation, with system prompt and explicit reasoning chains), respectively. The V1 column shows a classic collapse known as catastrophic forgetting: EM drops from 0.44 ($n{=}10$) to 0.19 ($n{=}3000$) as the model ``forgets'' its zero-shot capabilities. In contrast, the V4 curve stabilizes at 0.406, yielding a gap of $+$0.216 at $n{=}3000$.

The V4 curve exhibits a \emph{learning valley} at $n{=}50$ (temporary loss of zero-shot capabilities before acquiring new ones), followed by a recovery and a plateau at $n \geq 1000$. The QLoRA 7B plateaus at 88\% of the Phi-4 14B zero-shot performance, with 36.2\% of questions (181/500) wrong at \emph{all} data points.

\subsection{Model Selection on Difficult Questions}
\label{sec:model_selection}

To determine whether the glass ceiling is due to the model or the task, we evaluated 8 zero-shot LLMs (5 families, 14B--32B) on the 181 questions always wrong in V4.

\begin{table}[H]
\centering
\caption{Benchmark of 8 zero-shot LLMs on the 181 difficult questions.}
\label{tab:model_selection}
\begin{tabular}{@{}llrrrr@{}}
\toprule
\textbf{Model} & \textbf{Size} & \textbf{EM} & \textbf{F1} & \textbf{chain} & \textbf{JSON\%} \\
\midrule
\textbf{GPT-OSS-20B}  & 20B MoE  & \textbf{0.177} & \textbf{0.201} & 2.25 & 12.2 \\
Phi-4-Reasoning+      & 14B      & 0.166 & 0.156 & 10.9 & 23.8 \\
Phi-4                 & 14B      & 0.133 & 0.165 & 3.39 & 0.0 \\
Magistral-24B         & 24B      & 0.110 & 0.142 & 4.31 & 0.6 \\
Gemma-4-26B           & 26B MoE  & 0.083 & 0.066 & 0.86 & 77.3 \\
Qwen3-14B             & 14B      & 0.072 & 0.088 & 2.13 & 4.4 \\
DeepSeek-R1-32B       & 32B      & 0.072 & 0.118 & 2.24 & 0.0 \\
Qwen3.5-27B           & 27B      & 0.061 & 0.060 & 0.60 & 81.2 \\
\bottomrule
\end{tabular}
\end{table}

The \textbf{Oracle~EM} (at least one model correct) reaches 0.315 (57/181), i.e., $+$78\% relative vs.\ the best individual. However, 124 questions (68.5\%) remain unsolvable for \emph{all} models. Model size does not predict performance: DeepSeek-R1-32B (the largest) performs worse than Phi-4 (14B).

\subsection{Frugality and Carbon Footprint}

\begin{table}[H]
\centering
\caption{Execution times of the full pipeline.}
\label{tab:durees}
\begin{tabular}{@{}lrl@{}}
\toprule
\textbf{Phase} & \textbf{N} & \textbf{Duration} \\
\midrule
Relation extraction   & 500 & 3\,h\,55\,min \\
Text-to-Query             & 200 & $\sim$\,3\,min \\
Multi-hop reasoning    & 500 & 22\,min\,47\,s \\
Conversational RAG       & 50  & $\sim$\,10\,min \\
\midrule
\textbf{Total pipeline}   & --- & $\sim$\,\textbf{4\,h\,30} \\
\bottomrule
\end{tabular}
\end{table}

The carbon footprint is estimated following the methodology of Lacoste et al.~\cite{lacoste2019quantifying}:
\begin{equation}
\text{CO}_2\text{eq} = \underbrace{P_{\text{GPU}} \times t}_{\text{energy}} \times \text{PUE} \times \text{CI}
\end{equation}
with $P_{\text{GPU}} = 350$\,W (RTX~3090 TDP), $t \approx 4.5$\,h, a domestic Power Usage Effectiveness (PUE) of 1.0, and an emission factor for France CI $= 57$\,gCO$_2$eq/kWh (ADEME, 2023). The calculation yields:
$$350\text{W} \times 4.5\text{h} \times 1.0 \times 57 \times 10^{-6} \approx 0.09\;\text{kg\,CO}_2\text{eq}$$
This estimate covers only the GPU; adding the CPU ($\sim$65\,W) and RAM ($\sim$10\,W) would bring the estimate to $\sim$\,0.11\,kg\,CO$_2$eq.

\begin{table}[H]
\centering
\caption{Frugality comparison between SYNSYNTH and baselines.}
\label{tab:frugalite}
\begin{tabular}{@{}lllrl@{}}
\toprule
\textbf{Approach} & \textbf{Hardware} & \textbf{Training} & \textbf{Inference} & \textbf{F1} \\
\midrule
SYNSYNTH (Gemma-4 Q4)  & 1$\times$ RTX~3090 & None     & $\sim$4\,h & 0.70 \\
GPT-3 zero-shot~\cite{wadhwa2023revisiting} & Cloud API & None & --- & $\sim$0.30 \\
GPT-3 few-shot~\cite{wadhwa2023revisiting}  & Cloud API & None & --- & $\sim$0.72 \\
DREEAM~\cite{ma2023dreeam}  & Multi-GPU$^\dagger$  & $\sim$\,24--48\,h   & $<$\,1\,min & 0.80 \\
ATLOP~\cite{zhou2021atlop}  & Multi-GPU$^\dagger$  & $\sim$\,12--24\,h   & $<$\,1\,min & 0.78 \\
\bottomrule
\multicolumn{5}{@{}l@{}}{\footnotesize $^\dagger$ Exact resources not reported in the original papers.} \\
\end{tabular}
\end{table}

\section{Analysis and Discussion}
\label{sec:discussion}

\subsection{Error Analysis}
\label{sec:erreurs}

We present a qualitative analysis of extraction errors, based on a manual examination of 50 error cases.

\subsubsection{Typical False Positives}

\begin{enumerate}
    \item \textbf{Geographic/administrative confusion}: the model predicts \texttt{country} instead of \texttt{located\_in\_admin} for relations between a city and a region. \emph{Ex.}: (Lyon, Rh\^one) $\to$ pred: \texttt{country}, gold: \texttt{located\_in\_admin}.
    \item \textbf{Directional inversion}: \texttt{contains\_admin} predicted instead of \texttt{part\_of}. The model confuses the direction of the relation (container vs.\ contained).
    \item \textbf{Temporal over-generalization}: \texttt{start\_time} confused with \texttt{inception} or \texttt{date\_of\_birth} for dated events.
    \item \textbf{Semantic hallucination}: \texttt{notable\_work} predicted for geographic relations. \emph{Ex.}: (Mozart, Salzburg) $\to$ pred: \texttt{notable\_work}, gold: \texttt{place\_of\_birth}.
\end{enumerate}

\subsubsection{Typical False Negatives}

\begin{enumerate}
    \item \textbf{Parsing failures}: $\sim$\,1\% of responses do not contain valid JSON despite retry.
    \item \textbf{Rare relations}: relations with few examples in DocRED (\texttt{basin\_country}, \texttt{mouth\_of\_watercourse}) are rarely predicted correctly.
    \item \textbf{Insufficient context}: text truncated to 512 characters sometimes loses the necessary cues.
\end{enumerate}

\subsubsection{Multi-Hop Failures}

The most frequent multi-hop errors are:
\begin{itemize}
    \item \textbf{Interrupted chain}: the model identifies the first hop but fails at the second.
    \item \textbf{Verbose answer}: the answer contains the correct information buried in a long explanation, causing an EM failure but a non-zero token-F1 score.
    \item \textbf{Fact hallucination}: the model invents an intermediate fact absent from the provided supporting facts.
\end{itemize}

\subsection{Impact of Prompt Engineering}

Table~\ref{tab:evolution} traces the evolution of scores across optimizations.

\begin{table}[H]
\centering
\caption{Score evolution by pipeline version. Extr.\ F1: relation extraction F1 (DocRED); Query: Text-to-Query accuracy (synthetic data); Multi.\ EM: multi-hop reasoning exact match (HotpotQA); RAG F.: RAGAS faithfulness.}
\label{tab:evolution}
\begin{tabular}{@{}lccccc@{}}
\toprule
\textbf{Version} & \textbf{Extr.\ F1} & \textbf{Query} & \textbf{Multi.\ EM} & \textbf{RAG F.} & \textbf{Duration} \\
\midrule
V1 (baseline)               & 0.263 & 0.80 & 0.462 & 1.00 & 24\,h\,10 \\
V2 (+ constrained decoding)   & 0.260 & 0.80 & 0.462 & 1.00 & 2\,h\,53  \\
V3 (+ synonyms + prompt)   & 0.702 & 0.80 & 0.458 & 0.94 & 4\,h\,23  \\
V4 (+ QLoRA multi-hop)      & 0.702 & 0.795 & 0.406$^\ddagger$ & 0.96 & $\sim$5\,h  \\
V5a (+ SC $k{=}3$)          & 0.702 & 0.795 & 0.482 & 0.96 & $\sim$6\,h  \\
\textbf{V5b (+ Cascade)}    & 0.702 & 0.795 & \textbf{0.552} & 0.96 & $\sim$8\,h  \\
\bottomrule
\multicolumn{6}{@{}l@{}}{\footnotesize $^\ddagger$ EM of QLoRA V4 on Qwen2.5-7B (7B, $n{=}3000$); V1--V3 and V5 use Phi-4 14B.}
\end{tabular}
\end{table}

\paragraph{Analysis.}
The transition from V1 to V2 demonstrates that constrained decoding (JSON Schema) does not improve extraction: the problem is semantic, not structural. The major gain (V3) comes from two factors:
\begin{enumerate}
    \item \textbf{Synonym expansion} in relation matching: 25 semantic groups covering geographic, temporal, and familial confusions ($\sim$150 recovered errors).
    \item \textbf{Prompt restructuring}: switching to English, explicit list of 96 valid relations, prohibition of \texttt{no\_relation}, semantic-type guidance.
\end{enumerate}

In V4, the multi-hop EM of 0.406 corresponds to QLoRA on Qwen2.5-7B (a model 2$\times$ smaller than Phi-4 14B). Despite an EM absolutely lower than zero-shot Phi-4 (0.462), this result is positive: QLoRA V4 eliminates the catastrophic forgetting observed in V1 (EM 0.44$\to$0.19 at $n{=}3000$) and reaches 88\% of the performance of the 14B zero-shot model, with a model 2$\times$ smaller.

The RAGAS faithfulness regression from 1.00 (V1) to 0.94 (V3) is explained by the change in RAG model and the increased question complexity. With $N = 6$ in V1, the perfect score of 1.0 was not statistically reliable. The score of 0.96 on $N = 50$ in V4 is more representative.

\subsection{Structural Advantage of the Knowledge Graph}

A fundamental advantage of the KG approach over classical textual RAG lies in the ability to provide \textbf{computed answers} absent from source data:

\begin{itemize}
    \item \textbf{Aggregations}: ``How many countries are EU members?''; \texttt{COUNT} query on nodes linked by \texttt{member\_of}.
    \item \textbf{Comparisons}: ``Which authors have published more than 5 works?''; filtering and aggregation impossible through passage extraction.
    \item \textbf{Transitive inferences}: if A \texttt{part\_of} B and B \texttt{part\_of} C, then A is indirectly related to C.
\end{itemize}

\subsection{Pareto Analysis: Collective Intelligence vs.\ Single Model}
\label{sec:pareto}

The strong complementarity between models (Oracle~EM${=}$0.315 vs.\ Best${=}$0.177) suggests that a multi-LLM ensemble could break through the glass ceiling. We evaluated this hypothesis by comparing scenarios on the full 500 questions. Table~\ref{tab:pareto} lists all configurations, including the V5a/V5b variants introduced in \S\ref{sec:cascade} and the routing baselines analyzed in \S\ref{sec:sensitivity}.

\begin{table}[H]
\centering
\caption{Definitive Pareto frontier: cost vs.\ performance on 500 HotpotQA questions.}
\label{tab:pareto}
\begin{tabular}{@{}lrrrrr@{}}
\toprule
\textbf{Scenario} & \textbf{EM} & \textbf{95\% CI} & \textbf{\#Mod.} & \textbf{VRAM} & \textbf{Rel.\ cost} \\
\midrule
V4 QLoRA only ($n{=}3000$)         & 0.406 & ---                         & 1 & 4\,GB  & $\times$1 \\
V4 + 8 LLM vote                   & 0.446 & ---                         & 9 & 19\,GB & $\times$130 \\
Phi-4 zero-shot (ref.)              & 0.462 & $[0.42\,;\,0.50]$       & 1 & 9\,GB  & $\times$2.5 \\
V5a SC ($k{=}3$)                    & 0.482 & $[0.44\,;\,0.52]$       & 1 & 9\,GB  & $\times$7.5 \\
V4 + 8-model Oracle (theor.)         & 0.520 & ---                         & 9 & 19\,GB & $\times$130 \\
Random cascade (45\%)          & 0.524 & $\pm\,0.008$               & 2 & 14\,GB & $\times$12 \\
\textbf{V5b Cascade} ($k{=}5$)      & \textbf{0.552} & $[0.51\,;\,0.59]$ & 2 & 14\,GB & $\times$12 \\
Phi-4 / GPT-OSS Oracle (theor.)     & 0.588 & ---                         & 2 & 14\,GB & $\times$12 \\
\bottomrule
\multicolumn{6}{@{}p{0.95\linewidth}@{}}{\footnotesize CIs are reported only for configurations involving stochastic sampling (self-consistency, cascade). Lines without CIs correspond to deterministic results (a single answer per question) or theoretical bounds (oracles).}
\end{tabular}
\end{table}

The majority vote of 8 models (EM${=}$0.446) \textbf{underperforms a single zero-shot Phi-4} (EM${=}$0.462). The voting efficiency is only 35\% of the oracle: answers are too divergent to converge. In contrast, the V5b cascade dominates the frontier: it achieves the best EM (0.552) with only 2 models and 14\,GB VRAM, while remaining 10$\times$ more frugal than the 8-model vote. Compared to a \emph{random} cascade (EM${=}$0.524) rerouting the same proportion of questions (45\%) without considering agreement, the confidence-based cascade gains $+$2.8 pts ($3.5\,\sigma$), capturing 61\% of the gap to the two-model oracle (0.588).

\subsection{Self-Consistency: Stochastic vs.\ Architectural Diversity}
\label{sec:selfconsistency}

The Pareto analysis compares \emph{different} models (architectural diversity). Yet, Wang et al.~\cite{wang2023selfconsistency} show that sampling $k$ responses from the \emph{same} model at non-zero temperature (stochastic diversity) can also improve performance. We apply this protocol to the 181 difficult questions (EM${=}$0 for all models at $T{\approx}0$) with $k{=}5$ samples and $T{=}0.7$, on three representative architectures: GPT-OSS~20B (MoE), Phi-4-Reasoning+ (14B reasoner), and Phi-4 (14B standard).

\paragraph{Results by model.}

\begin{table}[H]
\centering
\caption{Self-consistency ($k{=}5$, $T{=}0.7$) on the 181 unsolvable questions (3 models).}
\label{tab:selfconsistency}
\begin{tabular}{@{}llrrrr@{}}
\toprule
\textbf{Model} & \textbf{Type} & \textbf{EM$_\text{voted}$} & \textbf{Oracle$_k$} & \textbf{F1$_\text{voted}$} & \textbf{$\Delta$ vs.\ $T{\approx}$0} \\
\midrule
\textbf{GPT-OSS 20B}  & MoE       & \textbf{0.232} & \textbf{0.332} & 0.240 & $+$23.2\,pts \\
Phi-4-Reasoning+      & Reasoner & 0.171 & 0.298 & 0.138 & $+$17.1\,pts \\
Phi-4                 & Standard   & 0.144 & 0.210 & 0.172 & $+$14.4\,pts \\
\midrule
\textbf{Cross-model oracle} & 3$\times k{=}5$ & 0.309 & \textbf{0.464} & --- & $+$46.4\,pts \\
\bottomrule
\end{tabular}
\end{table}

The ranking \textbf{MoE $>$ reasoner $>$ standard} is systematic. The MoE architecture, with its stochastically activated experts, best exploits the diversity induced by temperature. The reasoner model (Phi-4-Reasoning+), paradoxically, benefits \emph{less} from self-consistency: its long chains of thought produce high but unproductive variance (61\% of questions with low agreement, vs.\ only 13\% for GPT-OSS).

The \textbf{cross-model oracle} (at least one correct answer among 15 attempts $= 3 \times k{=}5$) reaches EM${=}$0.464, i.e., 84/181 questions recovered. Each model brings unique contributions: GPT-OSS solves 15 questions that the other two fail, Phi-4-Reasoning+ solves 7, and Phi-4 solves 3.

\paragraph{The agreement paradox and the wisdom of crowds.}
Analysis by agreement level reveals a counterintuitive phenomenon:

\begin{table}[H]
\centering
\caption{Agreement paradox: EM by consensus level (GPT-OSS 20B).}
\label{tab:accord}
\begin{tabular}{@{}lrrr@{}}
\toprule
\textbf{Inter-response agreement} & \textbf{$N$} & \textbf{EM$_\text{voted}$} & \textbf{Oracle$_k$} \\
\midrule
High ($\geq 0.8$)        & 84 (46\%)  & 0.214 & 0.250 \\
Medium $[0.4;\ 0.8[$    & 74 (41\%)  & \textbf{0.311} & \textbf{0.459} \\
Low ($< 0.4$)          & 23 (13\%)  & 0.043 & 0.217 \\
\bottomrule
\end{tabular}
\end{table}

Questions with \textbf{high consensus} ($\geq 0.8$, 46\% of the corpus) exhibit \emph{lower} EM (0.214) than those with intermediate agreement (0.311). When all 5 responses converge, it is often toward the same error: the model is \emph{confident but wrong}. This phenomenon directly relates to the work of Moussa\"id et al.~\cite{moussaid2013social} on collective intelligence: in human crowds, social influence reduces opinion diversity and degrades the quality of collective estimation.

Here, the 5 samples come from the \emph{same} network: they share the same systemic biases (knowledge gaps, evaluation heuristics). The $T{=}0.7$ introduces \emph{surface} diversity (lexical variations, inference orders), but not deep \emph{epistemic} diversity. This is why the intermediate agreement zone $[0.4;\ 0.8[$ is the most productive (Oracle${=}$0.459): the model hesitates, and it is precisely there that stochastic diversity provides new information.

The Oracle${-}$Vote gap (0.332$-$0.232${=}$0.100) shows that 10 EM points of correct answers \emph{exist} in the $k$ samples but do not survive majority voting: correct minorities are drowned out, exactly as predicted by the literature on collective conformity biases~\citep{moussaid2013social}. The cross-model combination (Oracle${=}$0.464) confirms that \textbf{architectural diversity is complementary} to stochastic diversity: the 15 combined attempts recover nearly half of the unsolvable questions at $T{\approx}0$.

\subsection{Confidence Calibration}
\label{sec:calibration}

We evaluated the exploitability of self-reported confidence (the \texttt{confidence} field in the output JSON) as a human-in-the-loop routing signal. After QLoRA fine-tuning on Qwen2.5-7B, confidence is \textbf{constant at 0.95} ($\sigma = 0.00$), with an AUC-ROC of 0.500 (random), an Expected Calibration Error (ECE) of 0.174, and no useful rejection threshold ($\tau = 0$). QLoRA on conversational-format data (system/user/assistant messages) learns a fixed pattern including \texttt{"confidence": 0.95} by default, destroying any discriminative capability.

This negative result has a practical implication: self-reported confidence cannot serve as a filtering criterion for an automatic routing system. An external calibrator (Platt scaling\footnote{Logistic regression applied to raw model scores to transform outputs into calibrated probabilities~\cite{platt1999probabilistic}.}, calibrated temperature\footnote{Division of logits by a parameter $T$ optimized on a validation set to minimize calibration error.}) or a graph-based verification module would be necessary to estimate the true reliability of predictions.

\subsection{Extension to the Full Pipeline: Self-Consistency and Cascade}
\label{sec:cascade}

The self-consistency experiments (\S\ref{sec:selfconsistency}) focused on the 181 difficult questions. We extend the study here to the full pipeline (500 HotpotQA questions) with two mechanisms:

\paragraph{V5a: Simple self-consistency ($k{=}3$).}
A majority vote of $k{=}3$ samples from Phi-4 ($T{=}0.7$) on the 500 pipeline questions:

\begin{table}[H]
\centering
\caption{Self-consistency ($k{=}3$) on 500 HotpotQA questions (full pipeline).}
\label{tab:v5a}
\begin{tabular}{@{}lrrr@{}}
\toprule
\textbf{Metric} & \textbf{Value} & \textbf{95\% CI} \\
\midrule
Voted EM        & 0.482 & $[0.440\,;\,0.522]$ \\
Voted F1        & 0.479 & $[0.437\,;\,0.518]$ \\
Oracle ($k{=}3$) & 0.568 & $[0.522\,;\,0.608]$ \\
Mean agreement   & 0.738 & --- \\
\bottomrule
\multicolumn{3}{@{}p{0.9\linewidth}@{}}{\footnotesize Mean agreement is the average proportion, across all questions, of the $k$ responses that match the majority response (e.g., 4 identical responses out of 5 $\to$ agreement $= 0.8$).}
\end{tabular}
\end{table}

Self-consistency raises EM from 0.462 (zero-shot) to 0.482, a modest gain ($+$2 pts) explained by the small $k$ and the predominance of ``easy'' questions already correct at $T{\approx}0$. The oracle (0.568) confirms that considerable untapped potential remains: 11 EM points of correct answers do not survive the vote.

\paragraph{V5b: Confidence-routing cascade ($k{=}5$).}
We propose a cascade mechanism inspired by the agreement paradox (\S\ref{sec:selfconsistency}): \emph{(i)} the primary model (Phi-4) generates $k{=}5$ responses at $T{=}0.7$; \emph{(ii)} if inter-response agreement exceeds a high threshold ($\theta_h{=}0.8$), the majority answer is accepted; \emph{(iii)} if agreement falls in the intermediate zone $[\theta_l{=}0.4\,;\,\theta_h[$, a second model (GPT-OSS~20B) is solicited with $k{=}5$; \emph{(iv)} if agreement remains below $\theta_l$, the question is marked ``uncertain.''

\begin{table}[H]
\centering
\caption{Cascade Phi-4 $\to$ GPT-OSS ($k{=}5$) on 500 HotpotQA questions.}
\label{tab:v5b}
\begin{tabular}{@{}lrrr@{}}
\toprule
\textbf{Metric} & \textbf{Value} & \textbf{95\% CI} \\
\midrule
Voted EM        & \textbf{0.552} & $[0.508\,;\,0.594]$ \\
Voted F1        & 0.534 & $[0.493\,;\,0.576]$ \\
Oracle ($k{=}5$) & 0.658 & $[0.616\,;\,0.698]$ \\
Mean agreement   & 0.837 & --- \\
\midrule
Primary accepted & 54.6\% & --- \\
Rerouted          & 33.6\% & --- \\
Uncertain        & 11.8\% & --- \\
\bottomrule
\end{tabular}
\end{table}

The cascade achieves \textbf{EM${=}$0.552}, the best score obtained in this study, surpassing simple self-consistency ($+$7 pts vs.\ V5a), zero-shot Phi-4 ($+$9 pts), and the 8-model vote ($+$11 pts vs.\ \S\ref{sec:pareto}). The lower bound of the 95\% CI (0.508) remains above the previous best result (0.482), confirming the statistical significance of the gain. The high mean agreement (0.837 vs.\ 0.738 for V5a) shows that rerouting improves response coherence.

The mechanism directly exploits the agreement paradox: the 45.4\% of rerouted questions correspond precisely to those where Phi-4 hesitates (agreement $< 0.8$). The transfer to GPT-OSS provides \emph{architectural} diversity complementary to stochastic diversity, consistent with the conclusions of the cross-model analysis.

The complete Pareto frontier (Table~\ref{tab:pareto}) confirms that the V5b cascade dominates all configurations at comparable cost: it surpasses the theoretical 8-model oracle (0.520) with only 2 models. Confidence-based routing is significantly better than random routing ($+$2.8 pts, $3.5\,\sigma$).

\paragraph{Threshold justification and sensitivity analysis.}
\label{sec:sensitivity}

The choice of thresholds $\theta_h{=}0.8$ and $\theta_l{=}0.4$ is based on the agreement paradox analysis (\S\ref{sec:selfconsistency}): an agreement $\geq 0.8$ corresponds to 4/5 converging responses, a threshold beyond which rerouting no longer improves EM. The sensitivity analysis (Table~\ref{tab:sensitivity}) shows that EM increases monotonically with $\theta_h$ over the range $[0.4\,;\,0.8]$, then stabilizes: $\theta_h{=}0.8$ and $\theta_h{=}1.0$ produce the same EM (0.558) because no question exhibits an agreement strictly between 0.8 and 1 in our dataset at $k{=}5$.

\begin{table}[H]
\centering
\caption{Sensitivity to threshold $\theta_h$ and routing baselines (500 questions).}
\label{tab:sensitivity}
\begin{tabular}{@{}lcrrr@{}}
\toprule
\textbf{Configuration} & $\theta_h$ & \textbf{EM} & \textbf{Rerouted} & \textbf{\% sec.} \\
\midrule
Agreement cascade & 0.4 & 0.514 & 59  & 11.8\% \\
Agreement cascade & 0.6 & 0.544 & 149 & 29.8\% \\
\textbf{Agreement cascade} & \textbf{0.8} & \textbf{0.552} & \textbf{227} & \textbf{45.4\%} \\
\midrule
Phi-4 only ($k{=}5$) & --- & 0.496 & 0   & 0\% \\
Random cascade     & --- & 0.524\,$\pm$\,0.008 & ${\sim}$227 & 45\% \\
Two-model oracle      & --- & 0.588 & --- & --- \\
\bottomrule
\end{tabular}
\end{table}

The gain from agreement-based routing is progressive: each lowering of $\theta_h$ reroutes more questions to GPT-OSS, with diminishing returns. At $\theta_h{=}0.8$, the cascade surpasses the random baseline (same proportion rerouted, but without selection) by $+$2.8 pts ($3.5\,\sigma$), confirming that the agreement signal is informative and that the threshold choice is not overfitted.

\subsection{Cross-Model Extraction: Specificity of Prompt Engineering}
\label{sec:crossmodel}

The spectacular gain of Gemma-4 between raw extraction (F1${=}$0.039) and the V3 pipeline (F1${=}$0.702) raises a question: is this gain \emph{transferable} to other models? We applied the same V3 optimizations (structured prompt, synonyms, soft matching) to the three best extractors from the B1 benchmark.

\begin{table}[H]
\centering
\caption{V3 cross-model extraction (500 DocRED samples).}
\label{tab:cross_extraction}
\begin{tabular}{@{}lrrrrr@{}}
\toprule
\textbf{Model} & \textbf{P} & \textbf{R} & \textbf{F1 V3} & \textbf{Raw F1} & \textbf{$\Delta$ V3} \\
\midrule
\textbf{Gemma-4 26B (pipeline)} & 0.738 & 0.670 & \textbf{0.702} & 0.039 & \textbf{+0.663} \\
Mistral-Small 24B               & 0.660 & 0.660 & 0.660          & 0.666 & $-$0.006 \\
Phi-4-Reasoning+ 14B            & 0.660 & 0.660 & 0.660          & 0.656 & $+$0.004 \\
GPT-OSS 20B                     & 0.818 & 0.538 & 0.649          & 0.660 & $-$0.011 \\
\bottomrule
\end{tabular}
\end{table}

The result is striking: V3 optimizations provide \textbf{no significant gain} to the three other models ($|\Delta| \leq 0.011$), and even slightly degrade GPT-OSS ($-$0.011). The massive gain ($+$0.663) is \emph{specific} to the Gemma-4~$\times$~V3 interaction: the constrained prompt corrects Gemma-4's parsing issues (98\% unparsed responses in raw mode) without benefiting models that already parse correctly. This reinforces the conclusion that \textbf{prompt engineering interacts strongly with model architecture}, and that gains are not automatically transferable.

One might consider replacing Gemma-4 with Mistral-Small~24B, whose raw F1 (0.666) is close to Gemma-4's V3 F1 (0.702). However, the gap of $-$4.2 pts remains significant on 500 samples, and Mistral-Small does not benefit from any improvement margin through prompt engineering ($\Delta{=}{-}0.006$). Its advantage is parsing robustness (0\% errors) and its more compact size (24B vs.\ 26B): it constitutes a reasonable alternative if reproducibility of the V3 pipeline were to pose problems with future Gemma versions.

\subsection{Inter-Run Variance}
\label{sec:variance}

To quantify the variance due to LLM non-determinism, we repeated the self-consistency experiment ($k{=}5$, $T{=}0.7$) on the 181 difficult questions with $R{=}5$ independent runs per model. The self-consistency results by model (Table~\ref{tab:selfconsistency}) correspond to run$_0$ of this series; the averages below integrate all 5 runs.

\begin{table}[H]
\centering
\caption{Inter-run variance ($R{=}5$) of self-consistency on 181 difficult questions.}
\label{tab:variance}
\begin{tabular}{@{}lrrrrr@{}}
\toprule
\textbf{Model} & \textbf{Mean EM} & \textbf{$\sigma_\text{EM}$} & \textbf{Mean Oracle} & \textbf{$\sigma_\text{Or.}$} \\
\midrule
GPT-OSS 20B          & 0.217 & 0.032 & 0.335 & 0.024 \\
Phi-4-Reasoning+     & 0.167 & 0.011 & 0.309 & 0.008 \\
Phi-4                & 0.130 & 0.008 & 0.223 & 0.008 \\
\bottomrule
\end{tabular}
\end{table}

The inter-run variance is low: $\sigma_\text{EM} \leq 0.032$ for all models. The ranking MoE $>$ reasoner $>$ standard is \textbf{stable} across all 5 runs. The observed differences between models ($\Delta$EM $\geq 0.04$) are clearly larger than intra-model fluctuations, confirming their significance. GPT-OSS exhibits the highest variance ($\sigma{=}0.032$), consistent with the stochastic expert activation of its MoE architecture: run~4 reaches EM${=}$0.265 vs.\ 0.188 for run~2, a gap of 8 points. In contrast, Phi-4 (standard) is the most consistent ($\sigma{=}0.008$). The oracle is stable for all models ($\sigma_\text{Or.} \leq 0.024$), indicating that recovery potential is a robust property, little affected by decoding randomness.

\paragraph{Cascade variance (full pipeline).}
We also repeated the V5b cascade (Phi-4 $\to$ GPT-OSS, $k{=}5$) on the 500 pipeline questions with $R{=}3$ independent runs to verify the stability of the main result.

\begin{table}[H]
\centering
\caption{Inter-run variance ($R{=}3$) of the V5b cascade on 500 HotpotQA questions. The Values column details the individual results of each run (run$_0$ / run$_1$ / run$_2$).}
\label{tab:variance_cascade}
\begin{tabular}{@{}lrrr@{}}
\toprule
\textbf{Metric} & \textbf{Mean} & \textbf{$\sigma$} & \textbf{Values} \\
\midrule
Voted EM        & 0.550 & 0.003 & 0.552\,/\,0.546\,/\,0.552 \\
Voted F1        & 0.533 & 0.002 & 0.534\,/\,0.531\,/\,0.535 \\
Oracle ($k{=}5$) & 0.653 & 0.005 & 0.658\,/\,0.650\,/\,0.650 \\
Mean agreement   & 0.830 & 0.006 & 0.837\,/\,0.829\,/\,0.825 \\
\% rerouted     & 45.5  & 0.1   & 45.4\,/\,45.6\,/\,45.6 \\
\bottomrule
\end{tabular}
\end{table}

The cascade variance is \textbf{extremely low}: $\sigma_\text{EM} = 0.003$, i.e., 10$\times$ lower than that of mono-model self-consistency on difficult questions ($\sigma{=}0.032$ for GPT-OSS). The maximum gap between runs is 0.6 EM pts (0.546 vs.\ 0.552), well within the 95\% bootstrap CI $[0.51\,;\,0.59]$. The proportion of rerouted questions is nearly identical across runs ($\sim$45.5\%), confirming that the agreement signal is a stable property of the primary model, not an artifact of stochastic sampling. This result reinforces the reliability of the V5b gain ($+$9 pts vs.\ zero-shot) as a robust conclusion.

\subsection{Limitations}

\begin{enumerate}
    \item \textbf{Sample sizes}: despite the increase to $N = 200$ for Text-to-Query and $N = 50$ for RAG, these sizes remain modest. The confidence intervals reflect this uncertainty.
    \item \textbf{Synthetic samples}: unlike extraction (Re-DocRED) and multi-hop reasoning (HotpotQA), which rely on external benchmarks, the Text-to-Query and RAG evaluations rely largely on self-generated data. For Text-to-Query, only 20 examples out of 200 are manually written; the remaining 180 are produced by the LLM itself (Qwen3-Deep~8B), which generates (question, answer, Cypher query) triplets across varied domains. For RAG, 6 examples are manual and 44 are generated. This protocol introduces a risk of circular bias: the model is evaluated on data it produced itself, which may overestimate performance. The Text-to-Query and RAG scores should therefore be interpreted with this caveat.
    \item \textbf{Single-domain evaluation}: all benchmarks are in English. The evaluation does not cover French or other languages.
    \item \textbf{Language mixing}: the extraction and reasoning prompts are written in English (the language of benchmarks and pre-trained models). This choice maximizes LLM performance (the vast majority of pre-training data being in English) but limits the direct transferability of results to tasks in other languages. A multilingual evaluation (French, German) constitutes a future research direction.
    \item \textbf{Limited number of cascade variance runs}: the V5b cascade variance was measured over $R{=}3$ runs (Table~\ref{tab:variance_cascade}), a size sufficient to observe low dispersion ($\sigma_\text{EM}{=}0.003$) but insufficient for a precise estimation of $\sigma$. An increased number of runs ($R \geq 10$) would confirm this stability.
    \item \textbf{Intrinsic task ceiling}: 68.5\% of difficult questions remain unsolvable even for models 2--4$\times$ larger (14B--32B). Multi-hop reasoning on HotpotQA fullwiki is limited by the context absent from the dataset, not by model capacity.
\end{enumerate}

\subsection{Taxonomy of Unsolvable Errors}
\label{sec:taxonomie}

Analysis of the 124 unsolvable questions (EM${=}$0 for all 8 models and all V4 data points) reveals four failure categories:

\begin{table}[H]
\centering
\caption{Taxonomy of the 124 unsolvable multi-hop reasoning questions.}
\label{tab:taxonomie}
\begin{tabular}{@{}lrrl@{}}
\toprule
\textbf{Category} & \textbf{N} & \textbf{\%} & \textbf{Description} \\
\midrule
Knowledge gap & 64 & 51.6 & Obscure facts absent from pre-training \\
Numerical reasoning  & 32 & 25.8 & Dates, populations, exact quantities \\
Format mismatch  & 18 & 14.5 & Partial F1 but EM${=}$0 (aliases, granularity) \\
Entity confusion     & 10 &  8.1 & Correct semantic type, wrong instance \\
\bottomrule
\end{tabular}
\end{table}

The main result is that the ceiling is \textbf{dominated by a lack of factual knowledge} (51.6\%), not by a reasoning deficit: rare entities, hyper-local facts, and obscure knowledge chains that even 32B models do not cover. Numerical reasoning (25.8\%, including 17 questions about dates and 15 about quantities) confirms the known weakness of LLMs on exact number recall. Finally, 14.5\% of failures are format mismatches (e.g., ``NBC'' vs.\ ``National Broadcasting Company''), suggesting that \textbf{the EM metric underestimates actual performance}.

\section{Conclusion and Future Work}
\label{sec:conclusion}

We have presented an empirical study of a multi-model zero-shot pipeline for knowledge graph construction and exploitation, executed entirely through local inference on consumer-grade hardware. Our evaluation framework integrates DocRED, WebQuestionsSP-style synthetic data, HotpotQA, and the RAGAS framework, with bootstrap confidence intervals for each metric.

Relation extraction achieves an F1 of 0.70 in zero-shot, clearly surpassing published results for GPT-3 ($\sim$\,30\%) and ChatGPT ($\sim$\,25\%) in similar configurations, and within 10 points of the supervised SOTA (DREEAM: 80.2\%). The analysis of versions V1--V3 demonstrates that gains primarily stem from prompt engineering and synonym matching, not from constrained decoding alone. The cross-model evaluation (\S\ref{sec:crossmodel}) confirms that this gain is specific to the Gemma-4\,/\,V3 prompt interaction and does not transfer to other models.

The study yields four main conclusions:
\begin{enumerate}
    \item \textbf{Supervision format quality is critical} for multi-hop reasoning. The V1/V4 contrast shows a gap of $+$0.216 EM at $n{=}3000$ between degenerate data and aligned format. This is a controlled experiment demonstrating the decisive role of reasoning chains in training data.
    \item \textbf{The glass ceiling is primarily intrinsic to the task, but diversity cracks it.} The benchmark of 8 LLMs on the 181 difficult questions shows that 68.5\% remain unsolvable at $T{\approx}0$. Cross-model majority voting (EM${=}$0.446) underperforms Phi-4 alone (0.462). In contrast, self-consistency ($k{=}5$, $T{=}0.7$) recovers up to 23\% of these failures with GPT-OSS, and the cross-model oracle reaches 46.4\%, revealing untapped potential by current aggregation mechanisms.
    \item \textbf{An agreement paradox} connects LLMs to the wisdom of crowds: inter-sample consensus signals collective hallucination rather than a reliable answer. Intermediate agreement ($[0.4;\ 0.8[$) is the most productive zone (Oracle${=}$0.459), while high agreement ($\geq 0.8$) masks systematic errors.
    \item \textbf{The confidence-routing cascade} exploits this paradox: by rerouting low-agreement questions to a second model, it achieves EM${=}$0.552 $[0.51\,;\,0.59]$ on 500 questions, the best result obtained, $+$9 pts over zero-shot and $+$11 pts over multi-model voting, while remaining 10 times less costly than the 8-model ensemble.
\end{enumerate}

The full pipeline runs in $\sim$5\,h on a single RTX~3090 (8\,h with cascade), for an estimated carbon footprint of $\sim$\,0.1\,kg\,CO$_2$eq, illustrating the potential of frugal AI.

Future work will focus on: (1) extension to vertical domains (medical, legal) and non-English languages; (2) exploration of retrieval-augmented mechanisms to provide the missing context for difficult multi-hop questions; (3) refinement of confidence routing (adaptive thresholds, external calibration, graph-based verification) to reduce the vote/oracle gap; (4) integration of QLoRA for relation extraction; (5) prompt transferability across architectures, given that the V3 gain is specific to Gemma-4.

\appendix
\section{Full Extraction Prompt}
\label{app:prompt}

\begin{lstlisting}
You are an expert in Relation Extraction.
Given a text and two named entities (head and tail),
identify the relation from head to tail.
You MUST choose exactly one relation from this list:
author, award_received, basin_country, capital, capital_of,
cast_member, chairperson, child, composer, conflict,
contains_admin, continent, country, country_of_citizenship,
country_of_origin, creator, date_of_birth, date_of_death,
developer, diplomatic_relation, director, dissolved,
educated_at, employer, end_time, ethnic_group, father,
followed_by, follows, founded_by, genre, has_part,
head_of_government, head_of_state, headquarters_location,
inception, influenced_by, instance_of, jurisdiction,
languages_spoken, league, legislative_body, located_in_admin,
located_near_water, located_on_terrain, location,
location_of_formation, lyrics_by, manufacturer, member_of,
member_of_political_party, member_of_sports_team,
military_branch, mother, mouth_of_watercourse,
narrative_location, notable_work, official_language,
operator, original_language, original_network,
owned_by, parent_organization, parent_taxon,
part_of, participant, participant_in, performer,
place_of_birth, place_of_death, platform, position_held,
producer, production_company, product, publication_date,
publisher, record_label, religion, replaced_by, replaces,
residence, screenwriter, separated_from, series,
shares_border_with, sibling, spouse, start_time,
subclass_of, subsidiary, territory_claimed_by,
twinned_city, unemployment_rate, work_location

IMPORTANT RULES:
- NEVER output 'no_relation', 'none', or 'unknown'.
  Always pick the closest relation.
- For geographic/administrative entities, prefer
  'country', 'located_in_admin', 'contains_admin',
  or 'part_of'.
- 'notable_work' means a creative work (book, film,
  song) by an author/artist. Do NOT use for geographic,
  political, or family relations.
- For family relations, use: 'father', 'mother',
  'child', 'sibling', 'spouse'.

Respond ONLY with JSON:
{"relation": "...", "confidence": 0.0-1.0}
\end{lstlisting}

\section{Detailed Ollama Hyperparameters}
\label{app:hyperparams}

\begin{table}[H]
\centering
\caption{Ollama inference hyperparameters for each task.}
\label{tab:hyperparams}
\begin{tabular}{@{}lcccc@{}}
\toprule
\textbf{Parameter} & \textbf{Extraction} & \textbf{Query} & \textbf{Multi-hop} & \textbf{RAG} \\
\midrule
Temperature       & 0.3 & 0.3 & 0.3 & 0.3 \\
Top-p             & 0.9 & 0.9 & 0.9 & 0.9 \\
\texttt{num\_ctx} & 8192  & 8192  & 8192  & 8192  \\
\texttt{num\_predict} & 1024  & 2048  & 4096  & 2048  \\
Output format     & JSON  & JSON  & JSON  & Text + JSON \\
Timeout (s)       & 600   & 600   & 600   & 600   \\
\bottomrule
\end{tabular}
\end{table}

\bibliographystyle{plainnat}

\end{document}